\documentclass[10pt,twocolumn,letterpaper]{article}

\usepackage[cvprfinal]{cvpr}    

\usepackage[dvipsnames]{xcolor}

\definecolor{cvprblue}{rgb}{0.21,0.49,0.74}
\usepackage[pagebackref,breaklinks,colorlinks,citecolor=cvprblue]{hyperref}
\usepackage{algorithm}
\usepackage{algpseudocode}
\usepackage{comment}
\usepackage{graphicx}
\usepackage{hyperref}

\makeatletter
\def\algbackskip{\hskip-\ALG@thistlm}
\makeatother

\def\paperID{123} 
\def\confName{CVPR}
\def\confYear{2024}

\title{MFTF: Mask-free Training-free Object Level Layout Control Diffusion Model}

\author{Shan Yang \\
Independent Researcher  \\
{\tt\small s.yang@alumni.duke.edu}
}

\begin{document}
\twocolumn[{
\maketitle
\begin{center}
    \centering
    \captionsetup{type=figure}
    \includegraphics[width=1\textwidth]{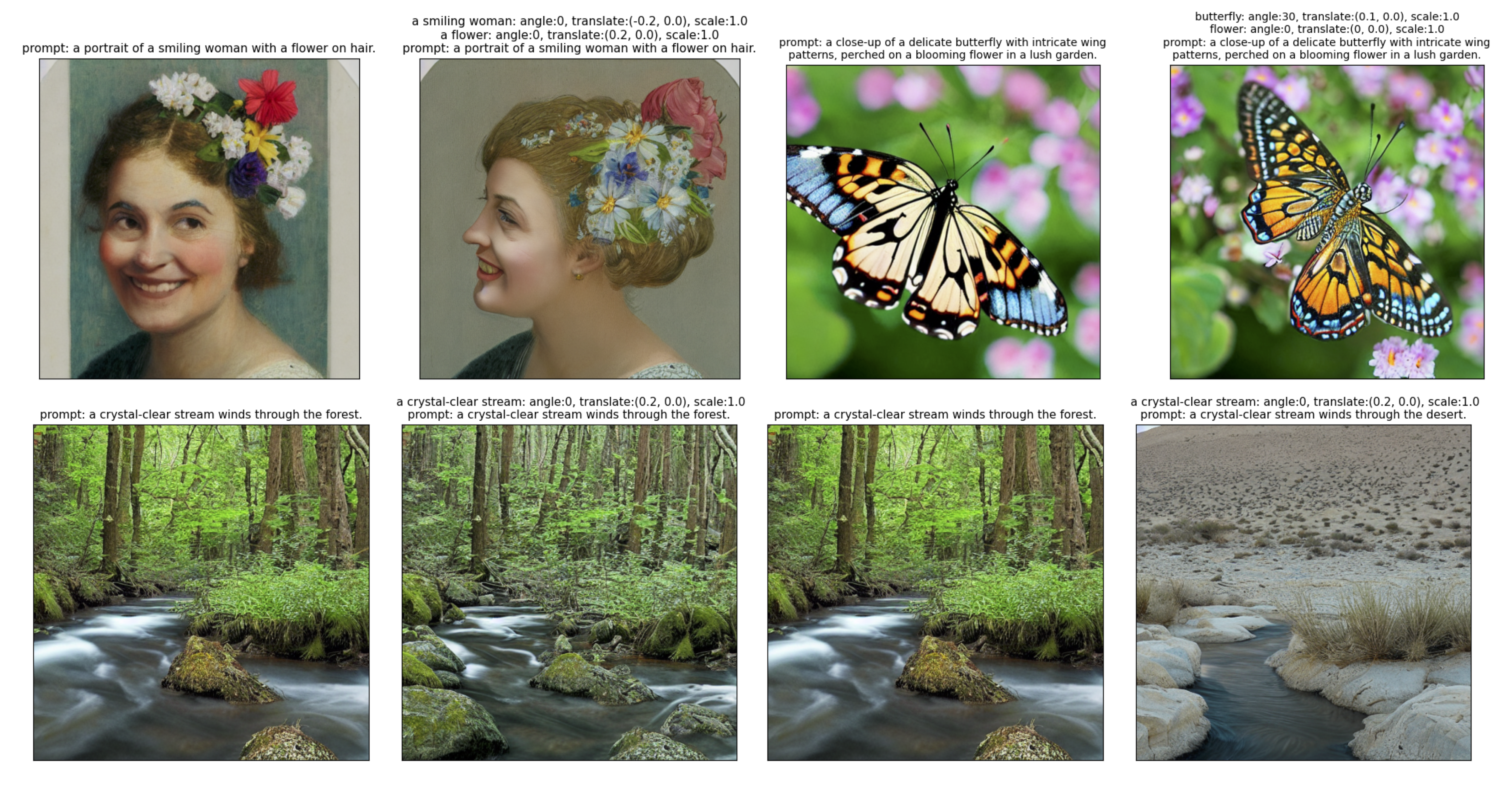}
    \captionof{figure}{MFTF successfully achieves single-object and multi-object layout control, as well as simultaneous layout control and semantic editing, without guidance image or mask and no model training and fine-tuning.}
    \label{fig:model-overview}
\end{center}
}]

\begin{abstract}
Text-to-image generation models have revolutionized content creation, but diffusion-based vision-language models still face challenges in precisely controlling the shape, appearance, and positional placement of objects in generated images using text guidance alone. Existing global image editing models rely on additional masks or images as guidance to achieve layout control, often requiring retraining of the model. While local object-editing models allow modifications to object shapes, they lack the capability to control object positions. To address these limitations, we propose the Mask-free Training-free Object-Level Layout Control Diffusion Model (MFTF), which provides precise control over object positions without requiring additional masks or images. The MFTF model supports both single-object and multi-object positional adjustments, such as translation and rotation, while enabling simultaneous layout control and object semantic editing. The MFTF model employs a parallel denoising process for both the source and target diffusion models. During this process, attention masks are dynamically generated from the cross-attention layers of the source diffusion model and applied to queries from the self-attention layers to isolate objects. These queries, generated in the source diffusion model, are then adjusted according to the layout control parameters and re-injected into the self-attention layers of the target diffusion model. This approach ensures accurate and precise positional control of objects. Project source code available at \href{https://github.com/syang-genai/MFTF}{https://github.com/syang-genai/MFTF}.
\end{abstract}
    
\section{Introduction}
\label{sec:intro}
Text-to-image generation models have become powerful tools for creating high-quality, photorealistic, and artistically refined images from natural language descriptions. Recent advances in vision-language generative modeling can be broadly categorized into two approaches: transformer-based autoregressive methods \cite{chen2020generative,yu2022scaling} and diffusion-based techniques \cite{sohl2015deep,ho2020denoising,song2020denoising,nichol2021improved}. Among these, diffusion models have demonstrated exceptional performance, achieving state-of-the-art results across a wide range of benchmarks \cite{nichol2021glide,rombach2022high,saharia2022photorealistic,ramesh2022hierarchical,betker2023improving}.

However, diffusion-based vision-language models struggle to provide precise control over the generated images using text guidance alone \cite{betker2023improving}. While images generated from the same text prompt with different initial noise typically maintain semantic consistency, they often display significant variations in object shape, appearance, and positioning \cite{patashnik2023localizing}.

Current layout control methods can be broadly categorized into global layout control and local object editing. Global layout control typically relies on additional masks or images as guidance, and may require model retraining depending on whether auxiliary networks are used \cite{tumanyan2023plug, zhang2023adding}. While existing local object editing techniques allow for control over semantics and structure, they still fall short of enabling precise control over the positional placement of objects \cite{hertz2022prompt, patashnik2023localizing, cao2023masactrl, kawar2023imagic}.

\begin{figure}
  \centering
   \includegraphics[width=1\linewidth]{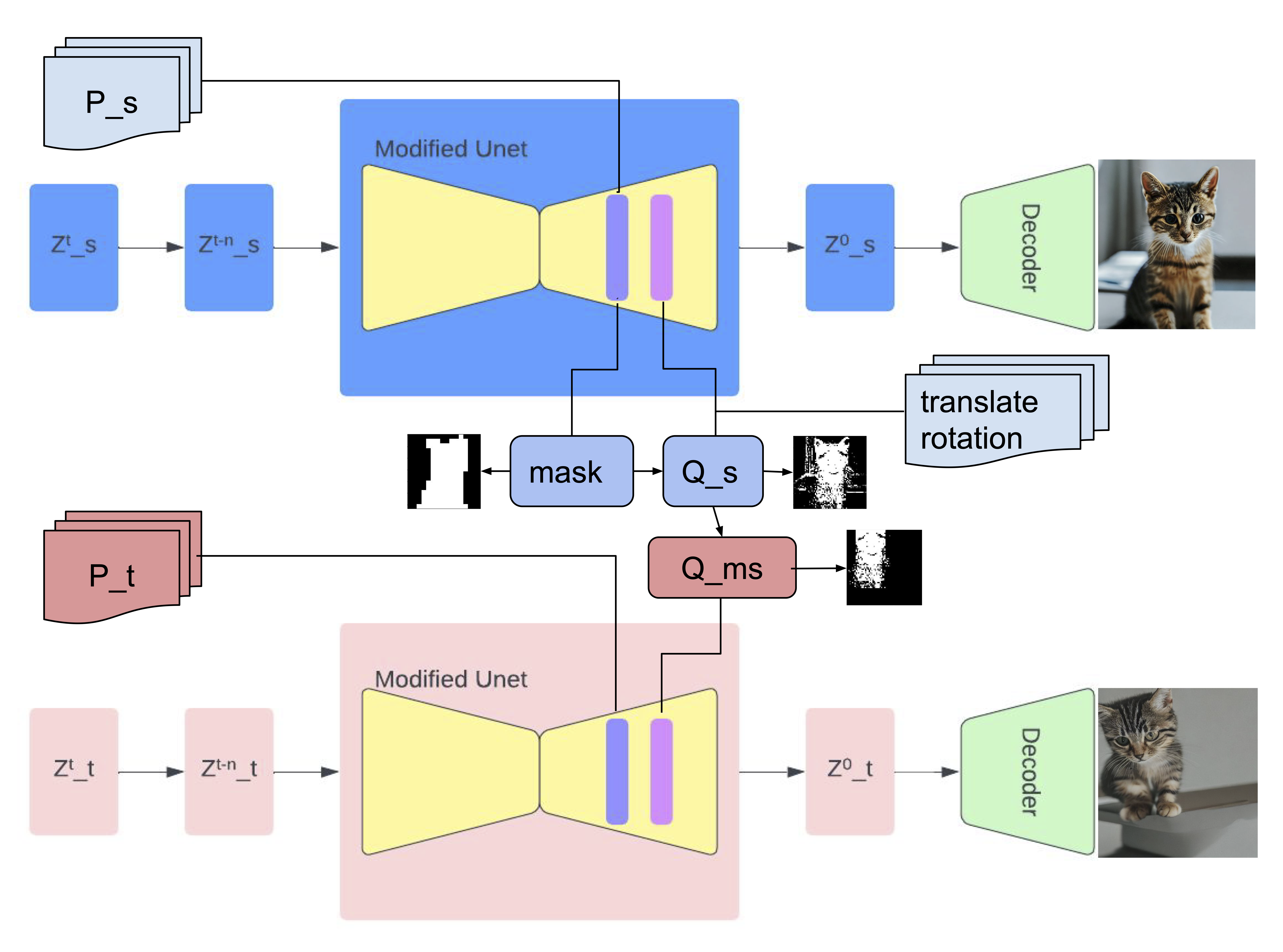}
   \caption{MFTF Architecture. MFTF dynamically generates attention masks from the cross-attention layers of the source diffusion model. These attention masks are then applied to the queries $Q_s$ derived from the self-attention layers to separate the objects from background. Subsequently, the modified queries $Q_{ms}$ are generated in accordance with the layout control parameters $L$. Finally, $Q_{ms}$ is injected into the self-attention layer of the target diffusion model, enabling precise positional control via denoising process.}
   \label{fig:architecture}
\end{figure}

To enhance existing spatial control techniques, we introduce MFTF model, which allows for precise positional control of individual objects without the need for additional masks or model retraining, as illustrated in \cref{fig:model-overview}. The architecture of the proposed model is shown in \cref{fig:architecture}. The model achieves this by manipulating the denoising process of the diffusion model through parallel denoising for both the source and target prompts. Attention masks are dynamically generated from the cross-attention layers of the source diffusion model and applied to queries from the self-attention layers to isolate objects. These modified queries are then adjusted based on layout control parameters and reintroduced into the self-attention layers of the target diffusion model, enabling accurate positional control throughout the denoising process. Furthermore, the cross-attention masks generated for tokens can be leveraged for text-guided semantic segmentation in S\cref{fig:segmentation}.

The challenges in achieving object-level layout control lie in doing so without relying on additional image guidance. Building on previous work \cite{hertz2022prompt, cao2023masactrl, tumanyan2023plug} that modifies either the cross-attention or self-attention layers and employs multi-self-attention techniques, our model dynamically generates attention masks. These masks are applied within the self-attention layers to effectively distinguish objects from the background. Ultimately, object layout control is accomplished through a modified multi-self-attention operation in supplement \cref{fig:model-arch}. The contributions of this model are as follows:
(1) The MFTF model enables single-object and multi-object level positional control, as well as concurrent object-level layout control and semantic editing. 
(2) MFTF supports concurrent layout control and semantic editing without requiring images or masks as guidance without model training or fine-tuning. 
(3) The proposed method can also be applied to text-controlled semantic segmentation.

\section{Related Work}
\label{sec:related}

\subsection{Text-to-Image Generation}
Text-to-image generation primarily encompasses two families of models: autoregressive models and diffusion models. Among the autoregressive models, the Parti \cite{yu2022parti} model integrates both text and image modules using a transformer architecture, enabling text-controlled, high-fidelity, and photorealistic image generation. Built upon VQVAE \cite{van2017neural} and VQGAN \cite{esser2021taming, yu2021vector}, the Parti model leverages these methods to tokenize and detokenize images, allowing for the fusion of text and image tokens within the transformer framework \cite{vaswani2017attention}.

The diffusion-based family of text-to-image generation models has garnered significant attention in recent years. Notable models in this category include Imagen \cite{saharia2022photorealistic}, which leverages pretrained large language models for text encoding, achieving strong text-image alignment and photorealistic image generation. OpenAI's DALL-E 2 and 3 \cite{ramesh2022hierarchical, betker2023improving} utilize the CLIP \cite{radford2021learning} model to align text and images, generating visuals via a diffusion model. The Latent Diffusion Model (LDM) employs a distinct architecture from DALL-E 2 and 3, conditioning on text for image generation and achieving text-image alignment within the latent space \cite{rombach2022high}. Our model builds upon the open-source Stable Diffusion model, which is an open-source implementation of the Latent Diffusion Model.

\begin{figure}
   \includegraphics[width=1\linewidth]{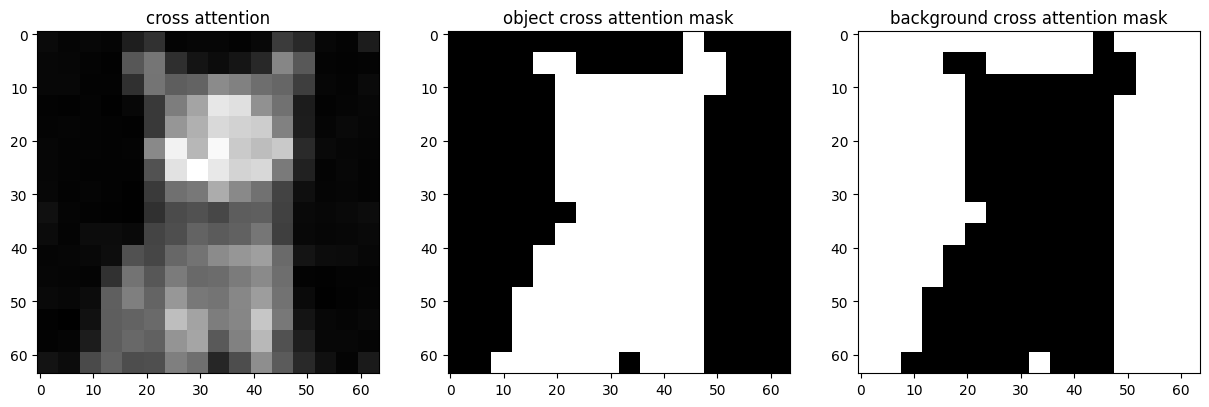}
   \caption{Cross-attention mask $M^i_s$ is generated from the cross-attention layers $A^i_s$}
   \label{fig:cross-attention-mask}
\end{figure}

\begin{figure}
   \includegraphics[width=1\linewidth]{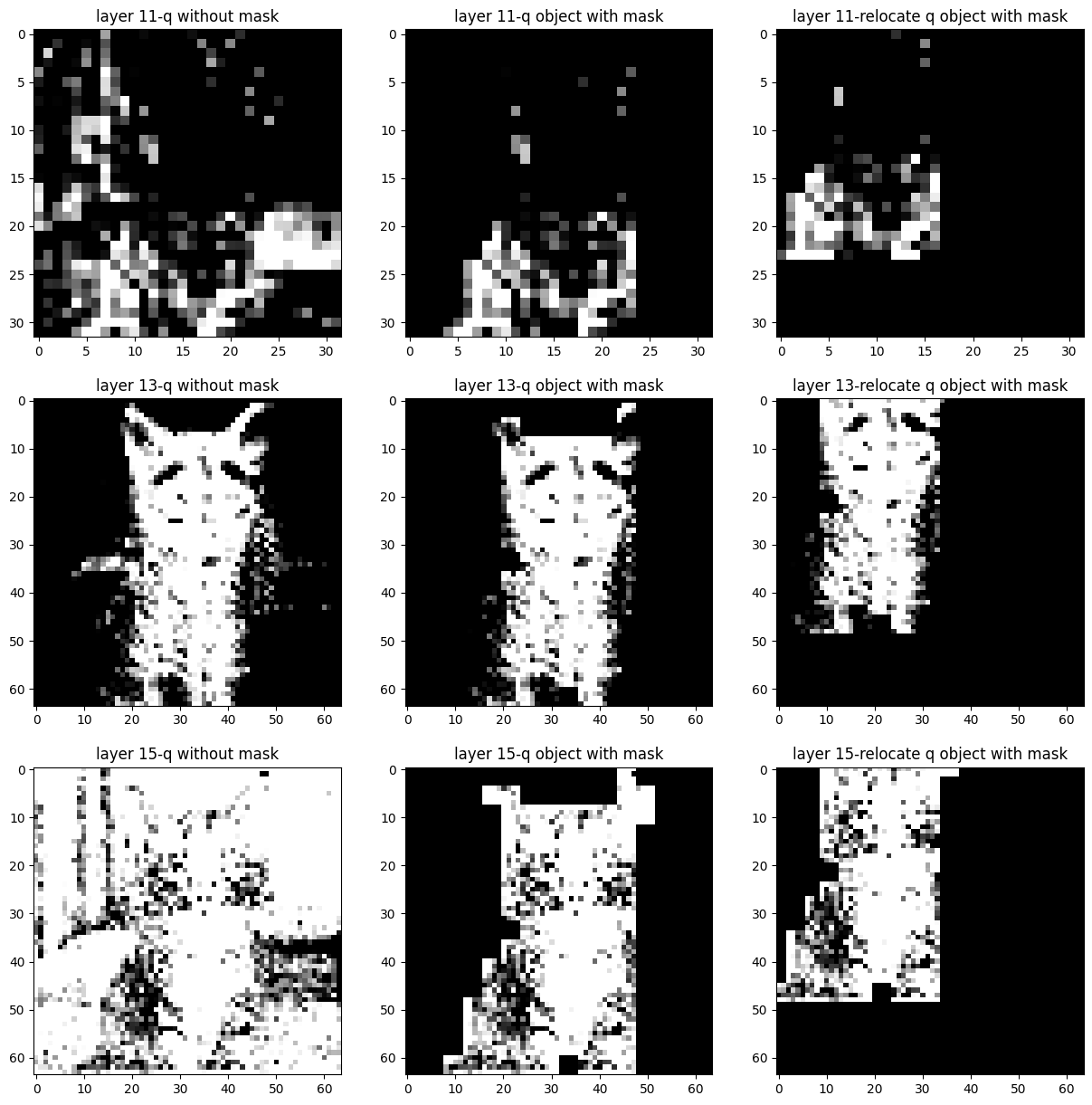}
   \caption{Visualization of $Q$ is presented at different self-attention layers $l = 11, 13, 15$, showing masks from the following conditions: without cross-attention mask, with cross-attention mask applied, and with additional positional control $L$.}
   \label{fig:self-attention-layers-10-15}
\end{figure}

\subsection{Text-to-Image Generation Layout Control and Semantic Editing}
While local object editing models enable shape modification of objects, they lack the ability to control the positional placement of objects. The Prompt-to-Prompt method \cite{hertz2022prompt} enables object editing while preserving the original object positions. Patashnik et al. \cite{patashnik2023localizing} applied prompt mixing combined with image-space localization techniques to edit the shape of local objects while maintaining the layout, semantics, and textures of the images. Masactrl \cite{cao2023masactrl} achieved object editing by modifying cross-attention and self-attention layers but still can not alter the positioning of objects. Imagic\cite{kawar2023imagic} performed local object editing through text embedding optimization and diffusion model fine-tuning. 

Global semantic editing and texture editing models that achieve layout control in diffusion models often rely on additional images or masks as guidance, with some requiring model training and others not. The Plug-and-Play model \cite{tumanyan2023plug} facilitates semantic editing while preserving the layout of the guidance image, which is achieved by injecting features into the self-attention layers. ControlNet \cite{zhang2023adding}, on the other hand, enhances layout control by incorporating "zero convolutions" and using external inputs such as instance edges, depth maps, segmentation, or human poses to manipulate the global image layout. However, ControlNet necessitates model training.

To address these limitations, our MFTF model enables precise control over the position of objects without requiring additional masks or images as guidance and no training. In addition to supporting single-object and multi-object positional control, including translation, rotation, etc.. It facilitates concurrent layout control and object semantic editing. MFTF model complements existing methods by enabling more flexible positional control in image generation. 

\section{Preliminaries}
\label{sec:preliminaries}
\subsection{Latent Diffusion Model}
Diffusion model is inspired by principles of non-equilibrium thermodynamics \cite{sohl2015deep, ho2020denoising, song2020denoising, nichol2021improved} and consists of two complementary processes: the forward process and the reverse process. In the forward process $q(x_t|x_{t-1})$, Gaussian noise $\epsilon$ is incrementally added to the original image $x_0$, progressively transforming it into random noise $x_T \sim N(0,I)$. In contrast, the reverse process $p_{\theta}(x_{t-1}|x_t)$ systematically removes the noise $\epsilon$ from the noisy image $x_t$, ultimately reconstructing the original image.

Our model builds upon the latent diffusion model \cite{rombach2022high}, which improves computational efficiency by performing the diffusion process in the latent space $z$. This approach significantly reduces the time complexity for forward and reverse processes while maintaining high-quality image synthesis \cite{rombach2022high}. Then latent diffusion model utilizes either a variational autoencoder (VAE) \cite{kingma2013auto, esser2020disentangling} or a VQGAN \cite{esser2021taming} for image encoding and decoding. Specifically, the encoder maps an input image $x$ into a compact latent representation $z$, and the decoder reconstructs $z$ back to the image domain $x$. The diffusion process, powered by a U-Net architecture \cite{ronneberger2015u}, operates entirely in the latent space $z$. To enable conditional guidance, a spatial transformer is integrated into the U-Net, providing enhanced flexibility. Additional conditions are encoded using an encoder $\tau_{\theta}$ and incorporated into the diffusion model through cross-attention layers within the spatial transformer. This architecture ensures robust and effective guidance throughout the diffusion process, facilitating high-quality and controlled image generation.

\subsection{Cross-attention and Self-attention in Diffusion Models}
In Stable Diffusion model, the modified U-Net module $\epsilon_{\theta}$ integrates a spatial transformer block, which includes a self-attention layer followed by a cross-attention layer. The mathematical formulations for self-attention and cross-attention are provided in \cref{self-attention} and \cref{cross-attention}, respectively. In these formulations, $Q_z$, $K_z$, and $V_z$ are computed from spatial features in \cref{self-attention}, $K_\tau$ and $V_\tau$ are derived from text embeddings in \cref{cross-attention}, $d$ denotes the embedding dimension.

\begin{equation}
\begin{aligned}
\text{self-attention}(Q_z,K_z,V_z)=\text{softmax}(\frac{Q_zK_z^T}{\sqrt{d}})*V_z
\end{aligned}
\label{self-attention}
\end{equation}

\begin{equation}
\begin{aligned}
\text{cross-attention}(Q_z,K_\tau,V_\tau)=\text{softmax}(\frac{Q_zK_\tau^T}{\sqrt{d}})*V_\tau
\end{aligned}
\label{cross-attention}
\end{equation}

The self-attention and cross-attention layers capture rich semantic and structural information, enabling their use for controlling style and semantic concepts \cite{hertz2022prompt,tumanyan2023plug}. Additionally, the cross-attention mechanism can generate masks to separate objects from the background, as employed in Masactrl \cite{cao2023masactrl}. MFTF applies the cross-attention mask $M^l_i$, objects corresponding to token $i$ can be effectively isolated from the background in self-attention layers. 

\begin{figure*}
  \centering
    \begin{subfigure}{0.495\linewidth}
	    \includegraphics[width=\linewidth]{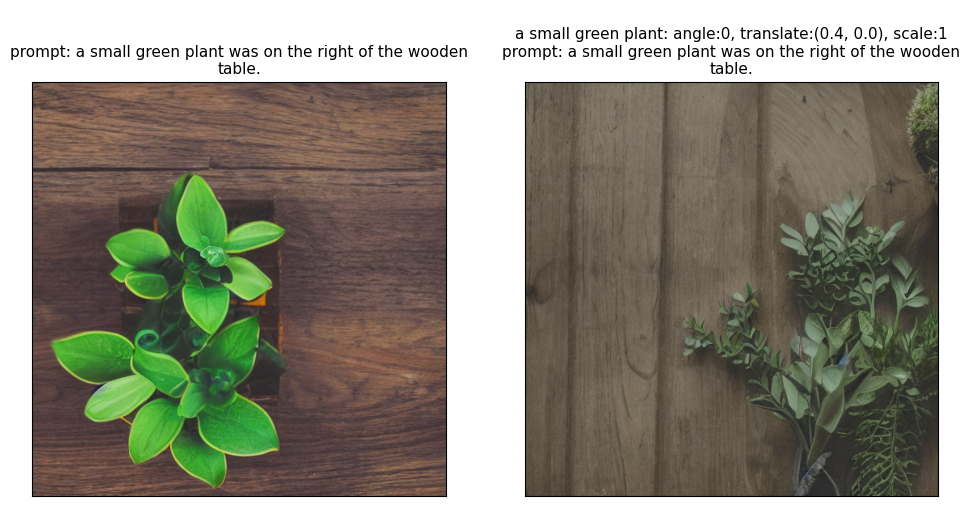}
	\end{subfigure}
	\begin{subfigure}{0.495\linewidth}
	    \includegraphics[width=\linewidth]{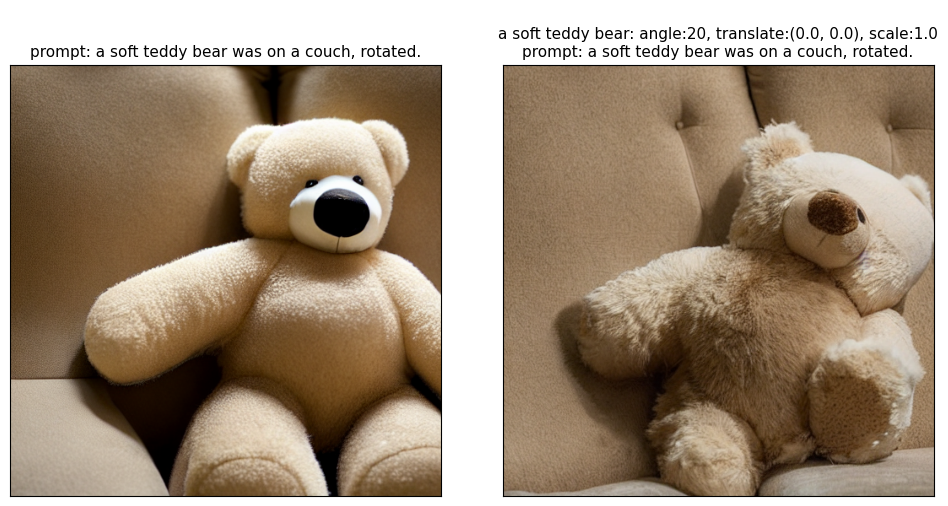}
	\end{subfigure}
    
    \vfill
	\begin{subfigure}{0.495\linewidth}
	     \includegraphics[width=\linewidth]{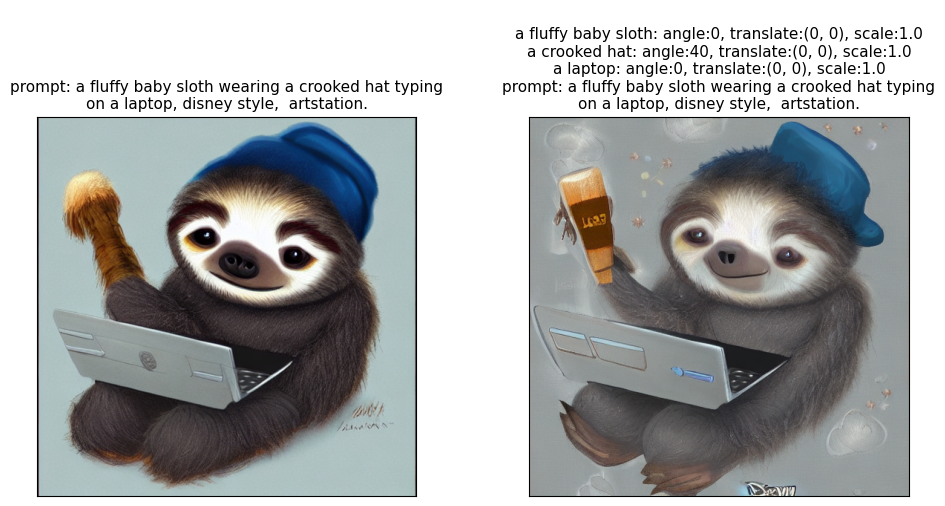}
	\end{subfigure}
	\begin{subfigure}{0.495\linewidth}
	    \includegraphics[width=\linewidth ]{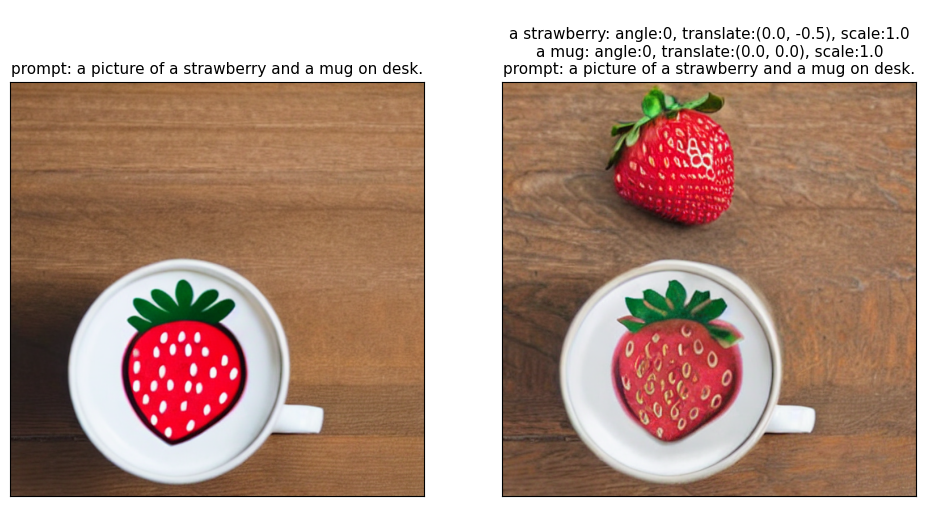}
    \end{subfigure}
    
    \vfill
	\begin{subfigure}{0.495\linewidth}
	     \includegraphics[width=\linewidth]{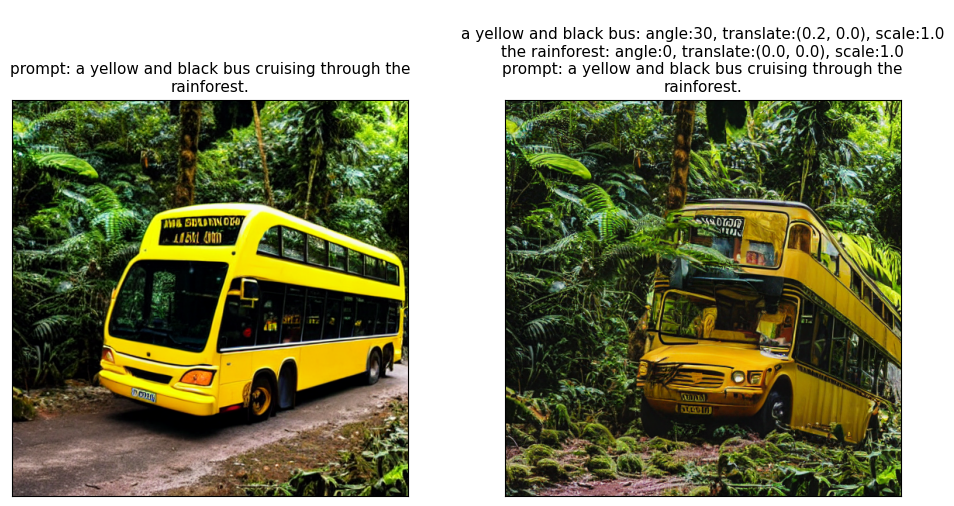}
	\end{subfigure}
	\begin{subfigure}{0.495\linewidth}
	    \includegraphics[width=\linewidth ]{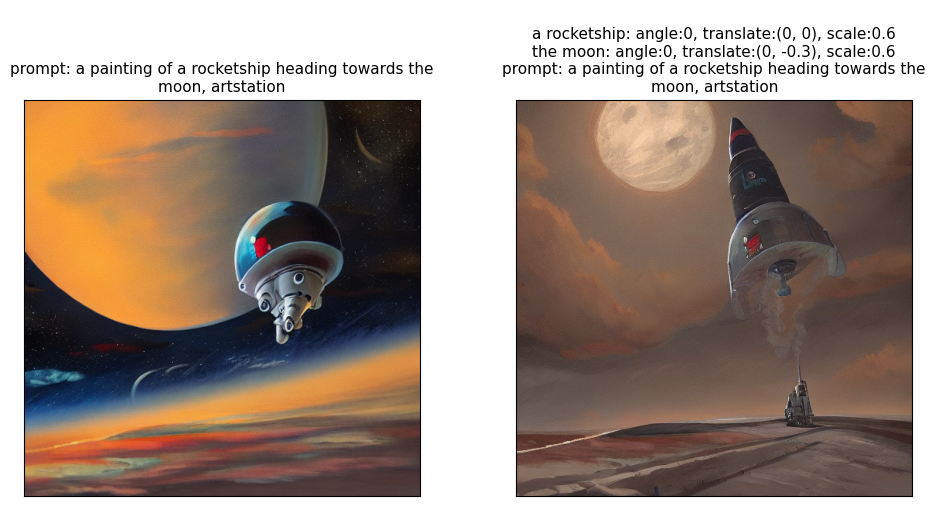}
    \end{subfigure}
    
    \vfill
    \begin{subfigure}{0.495\linewidth}
	    \includegraphics[width=\linewidth ]{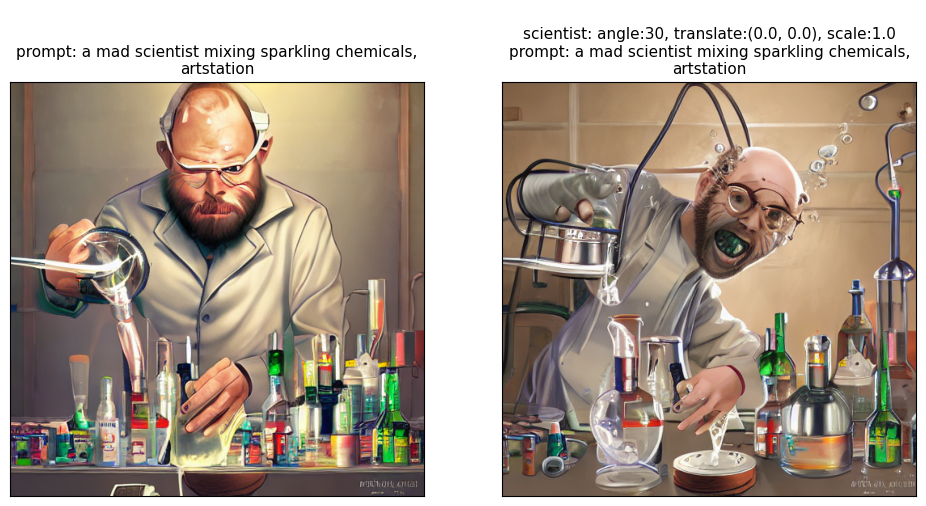}
    \end{subfigure}
    \begin{subfigure}{0.495\linewidth}
	    \includegraphics[width=\linewidth ]{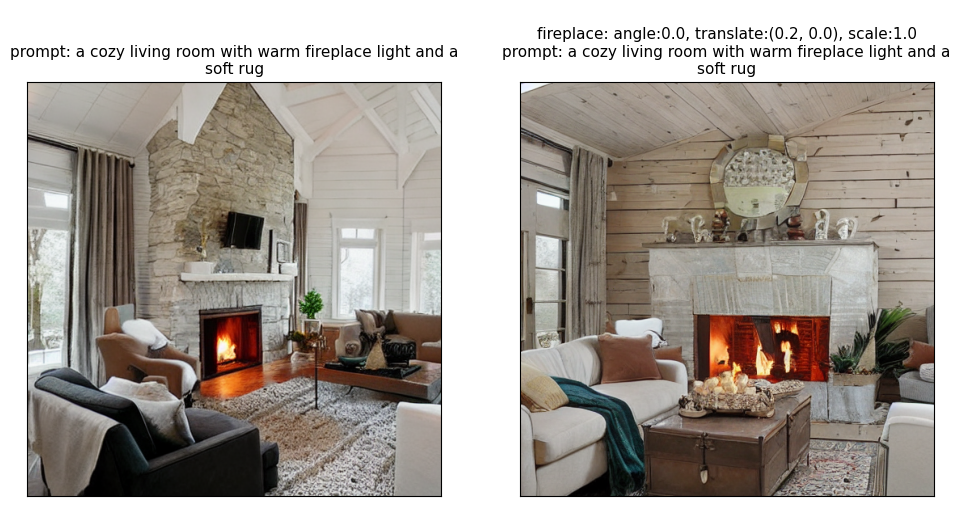}
    \end{subfigure}

   \caption{Quantitative comparison of images: objective-level layout control via text descriptions vs. MFTF model; without vs. with MFTF model.}
   \label{fig:experiment}
\end{figure*}

\section{Simultaneous Object Level Layout Control and Semantic Editing}
\label{sec:Simultaneous Object-Level Layout Control and Semantic Editing}
The primary aim of this model is to enable simultaneous object-level layout control and semantic editing without relying on additional image or mask guidance, as well as without necessitating model training or fine-tuning. Our MFTF model accomplishes this by dynamically generating attention masks and leveraging cross-attention mechanisms to separate objects from the background in the self-attention layers. Object layout control is subsequently realized through the application of modified multi-self-attention operations S\cref{fig:model-arch}. As illustrated in \cref{fig:model-overview}, MFTF model achieves precise positional control of individual objects, allowing layout transformations such as translation, scaling, and rotation to be applied concurrently. Furthermore, the model demonstrates the capability to perform object-level layout control and semantic editing simultaneously while also supporting the concurrent manipulation of multiple object positions.

The architecture of the proposed model is depicted in \cref{fig:architecture}. The model operates using a source prompt $P_s$ to generate source image $x_s$ and a target prompt $P_t$ to generate target image $x_t$. The target prompt $P_t$ can either be identical to $P_s$ or differ, depending on whether semantic editing is applied. To control the layout of objects in the generated target image $x_t$, the model requires the object tokens $P^i_s$ from $P_s$ and the object layout control parameters $L^i$. For instance, as illustrated in \cref{fig:model-overview}, the model adjusts the position of the "a crystal-clear stream" to the center based on $L$, and then replacing the "forest" in $P_s$ with "desert" in $P_t$. 

MFTF algorithm is summarized in \cref{alg:MFTF}. The parameter $t^*$ represents the cutoff time step for applying layout control, while $T$ denotes the total number of denoising time steps. For time steps $t < t^*$, the mask $M^i_s$ is obtained from the cross-attention layer, with $Q_s$ derived from the self-attention layer. The cross-attention mask $M^i_s$ is then applied to the query $Q_s$ in the self-attention to separate the object from the background.  Details of the cross-mask generation are provided in \cref{sec:cross-attention}. The object's position is adjusted according to the given layout control parameters $L$. Subsequently, the modified query $Q_{ms}$ is passed to the target diffusion model to generate the target latent space $z_t$. Details of the self-attention query $Q_{ms}$ generation are further explained in \cref{sec:self-attention}.

MFTF model enables simultaneous object-level layout control and semantic editing and through three key mechanisms: (1) the cross-attention mask contains semantic and structural information of tokens, despite potential overlaps in semantic features (S\cref{fig:cross-attention-map}); (2) query $Q$ from the self-attention layers incorporates layout information and structural details of the generated images, allowing positional control by modifying the objects within $Q_s$ (3) the diffusion model generate structural details during the later stages of the denoising process.  

\begin{algorithm}[h!]
\caption{MFTF: Mask-free Training-free Object Level Layout Control Diffusion Model} 
\label{alg:MFTF}
\textbf{Input}: \\
source: source prompt $P_s$, latent noise $z^T_s$, token index $i$, cross attention $A_s$, layout control parameters $L$, threshold $\eta$ \\
target: target prompt $P_t$, latent noise $z^T_t$ \\
other: $t^*$ cut off time-step for applying layout control, $T$ denoising time-step \\
\textbf{Output}: $z^0_s$ and $z^0_t$
\begin{algorithmic}
\State $t \gets T$
\While{$t>0$}
\If {$T-t>t^*$}
    \State $ \epsilon_s, (Q_s,K_s,V_s), A_s \gets \epsilon^s_\theta(z^t_s,P_s,t) $
    \State $ z^{t-1}_s \gets sample (z^t_s, \epsilon_s )$
    \State $ M_s \gets create-mask (A_s,\eta,i) $
    \State $ Q_{ms} \gets edit (M_s,Q_s,L) $
    \State $ \epsilon_t, (Q_t,K_s,V_s) \gets \epsilon^t_\theta(z^t_t,P_t,t,(Q_{ms},K_t,V_t))$
    \State $ z^{t-1}_t \gets sample (z^t_t, \epsilon_t) $
\Else{}
    \State $ \epsilon_s, (Q_s,K_s,V_s) \gets \epsilon^s_\theta(z^t_s,P_s,t) $
    \State $ z^{t-1}_s \gets sample (z^t_s, \epsilon_s )$
    \State $ \epsilon_t, (Q_t,K_s,V_s) \gets \epsilon^t_\theta(z^t_t,P_t,t)$
    \State $ z^{t-1}_t \gets sample (z^t_t, \epsilon_t) $
\EndIf
    \State $ t \gets t-1 $
\EndWhile
\end{algorithmic}
\end{algorithm}

\subsection{Cross-Attention Mask Creation}
\label{sec:cross-attention}
The cross-attention mask is constructed utilizing the cross-attention layer, where the object mask is derived following the formula described in \cref{eq:cross-attention-mask}, with a specified threshold $\eta$. The generated masks, containing both the object and the background, are depicted in \cref{fig:cross-attention-mask}. The threshold $\eta$ has a significant impact on the generated images through the attention mask. When $\eta = 0$, no attention mask is applied to the query, making it impossible to separate the object from the background. When $\eta = 1$, the query is entirely removed from the self-attention process. For multiple-object position control, the threshold $\eta$ controls the degree to which the original object is retained. 

\begin{equation}
\begin{aligned}
M^i=\begin{cases}
    1, & \text{if $A^i>=\eta$}\\
    0, & \text{otherwise} 
  \end{cases}
\end{aligned}
\label{eq:cross-attention-mask}
\end{equation}

The model dynamically generates cross-attention masks $M^i_s$ for token $i$ at each sampling step $t$ by leveraging the definition of the cross-attention map. The cross-attention map, represented as $A^i_s = QK^T$, has dimensions $b, n, m$, where $b$ corresponds to the batch size, $n$ is the feature sequence length, and $m$ is the text sequence length corresponds to token $0-m$. This enables controlled image generation by updating the cross-attention masks during each sampling step, ensuring layout control throughout the generation process.

The cross-attention mask primarily encodes the overall profile information of objects, lacking fine-grained structural details. In contrast, the self-attention layer captures detailed structural features and is instrumental in controlling both the precise location and fine-grained structure of objects within the generated image.  

\subsection{Self-Attention-Based Layout Control}
\label{sec:self-attention}
\begin{figure}[h!]
   \includegraphics[width=1\linewidth]{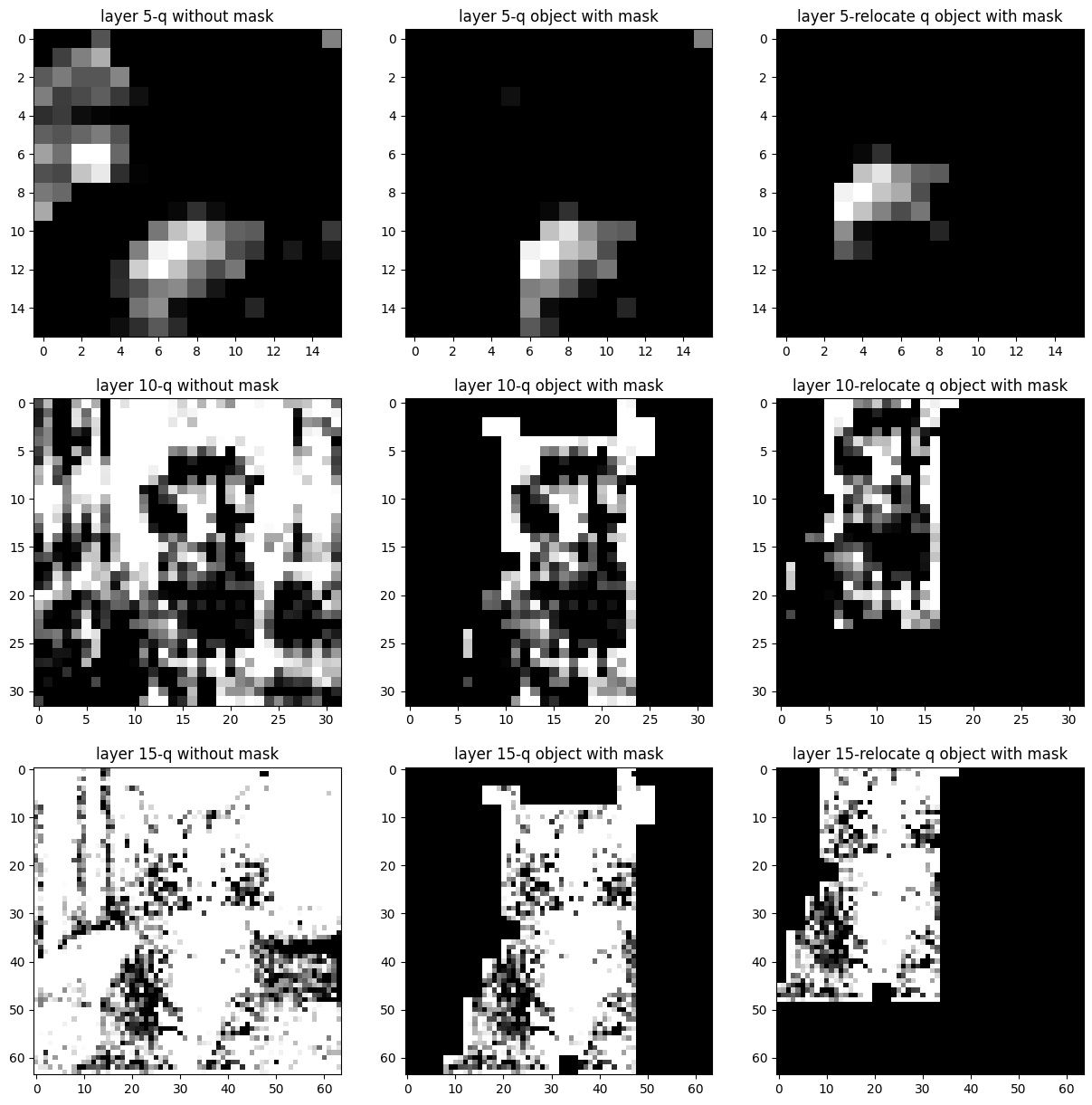}
   \caption{Visualization of $Q$ is presented at different self-attention layers $l=5,10,15$, showing masks from the following conditions: without cross-attention mask, with cross-attention mask applied, and with additional positional control $L$.}
   \label{fig:self-attention-layers-5-15}
\end{figure}

The self-attention layer $Q$ encodes both detailed object information and global layout characteristics of the generated images. The semantic and structural information represented by $Q$ varies across different self-attention layers. Lower layers ($l = 0$ to $5$) lack significant semantic information but begin to establish the foundational structure. In the intermediate layers ($l = 6$ to $10$), structural features become more prominent and start to take shape. The upper layers ($l = 11$ to $15$) contain detailed object-specific structural and semantic information. $Q_{s,l}$ corresponds to the self-attention layer $l$ as illustrated in S\cref{fig:self-attention-layers-5-15}.

The cross-attention mask enables effective objects selection from the background following \cref{eq:q_mask}, and is shown in the second column of \cref{fig:self-attention-layers-10-15}. 

\begin{equation}
\begin{aligned}
Q_{ms}=Q_s*M^i_s
\end{aligned}
\label{eq:q_mask}
\end{equation}

The layout of these objects is then controlled using the layout control parameters $L$ following \cref{eq:q_l}, allowing the manipulation of object locations in $Q_s$. This process generates the modified query $Q_{ms}$, as illustrated in the third column of \cref{fig:self-attention-layers-10-15}.
\begin{equation}
\begin{aligned}
Q_{ms}=l(Q_{ms},L)
\end{aligned}
\label{eq:q_l}
\end{equation} 

\begin{figure}[h!]
   \includegraphics[width=1\linewidth]{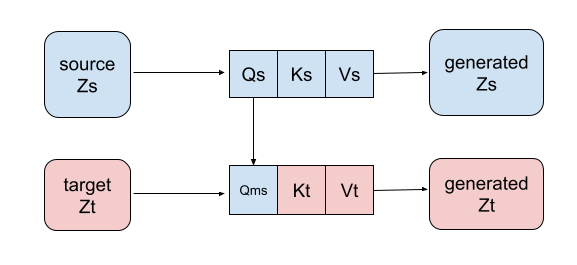}
   \caption{Modified multi-self-attention.}
   \label{fig:model-arch}
\end{figure}

In contrast to the multi-self-attention mechanism in Masactrl \cite{cao2023masactrl}, which utilizes $K_s$ and $V_s$ for the source diffusion model, MFTF model replaces $Q_t$ with $Q_{ms}$ in the target diffusion model to calculate self-attention, as illustrated in S\cref{fig:model-arch}. 

\subsection{Experiments}
\subsubsection{Experimental Setup}
We implement the proposed method using the state-of-the-art Stable Diffusion text-to-image model \cite{rombach2022high}, leveraging the publicly available v1.4 checkpoints. Sampling is performed using DDIM \cite{song2020denoising} with $30$ steps in total, and the classifier guidance scale set to $7.5$. The Gaussian random noise is initialized as $Z^T_s=Z^T_t$ for the source and target diffusion models Layout control is applied during steps $0\sim15,20$ across layers $0\sim16$. The threshold for mask generation ranges from $0.1$ to $0.3$ or is set to $1$ for objects to be dropped. 

\begin{figure*}
    \centering
    \begin{subfigure}{0.49\linewidth}
	    \includegraphics[width=\linewidth]{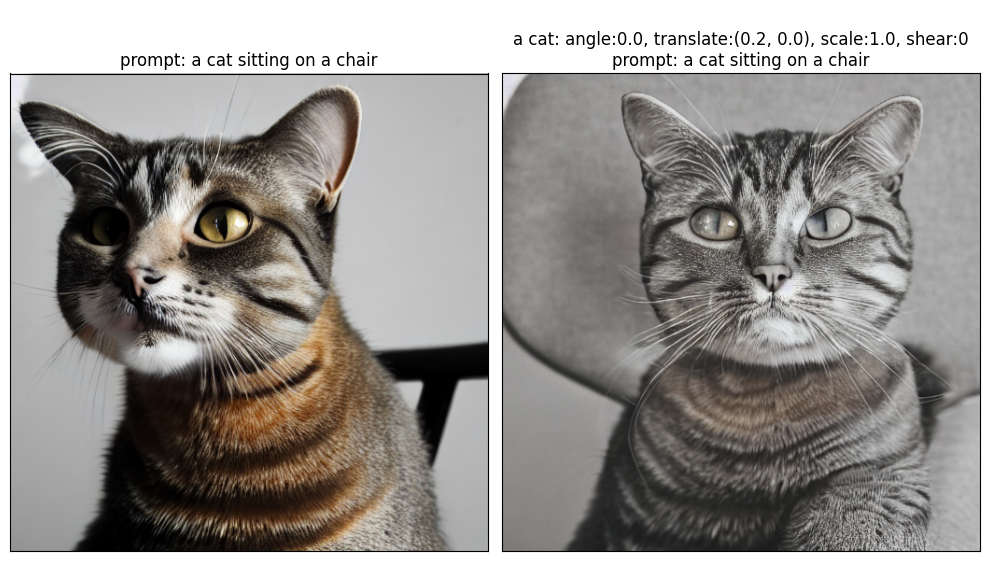}
	\end{subfigure}
	\begin{subfigure}{0.245\linewidth}
	     \includegraphics[trim=125mm 0 0 0, clip, width=\linewidth]{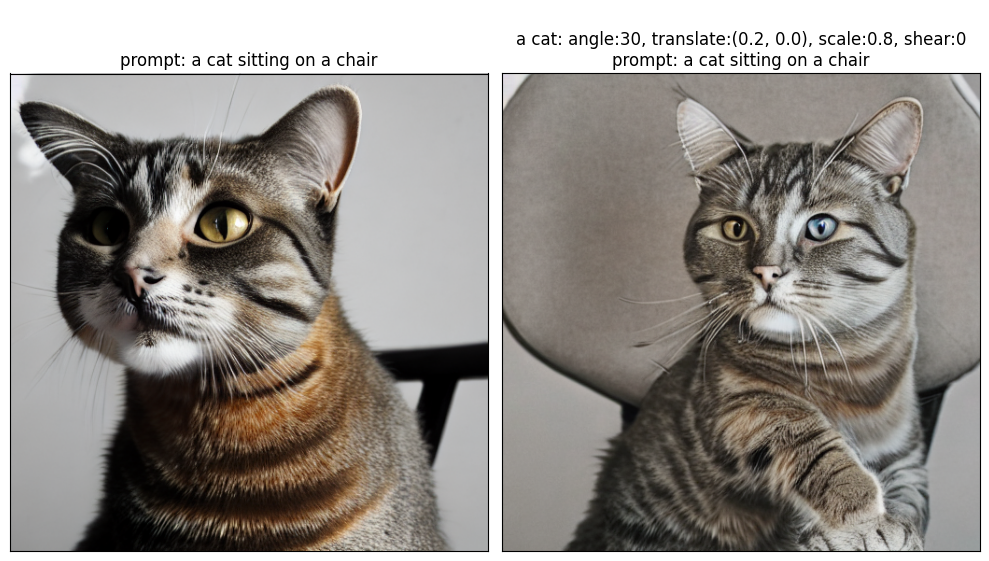}
	\end{subfigure}
    \begin{subfigure}{0.245\linewidth}
	     \includegraphics[trim=125mm 0 0 0, clip, width=\linewidth]{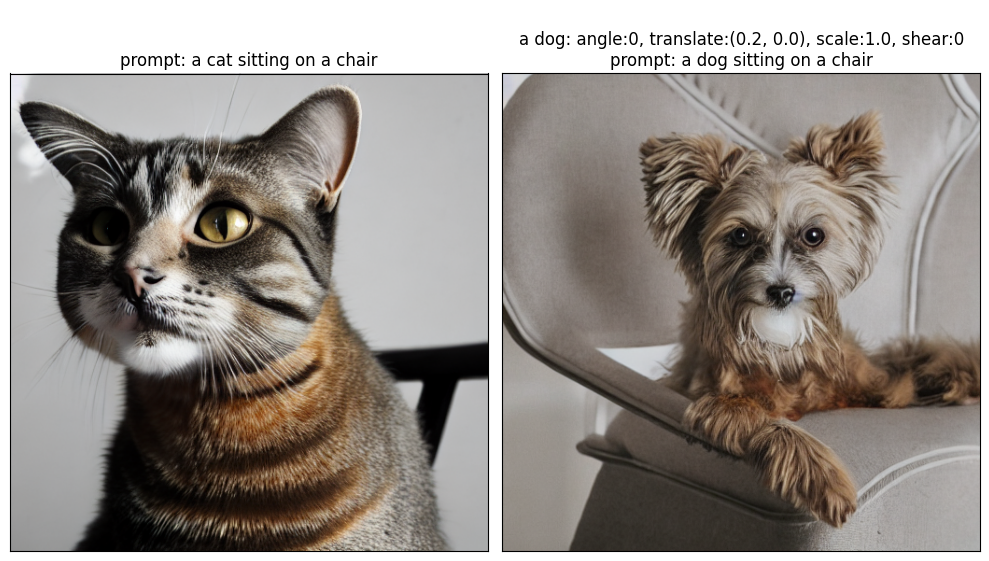}
	\end{subfigure}
    
    \vfill
    \begin{subfigure}{0.49\linewidth}
	    \includegraphics[width=\linewidth ]{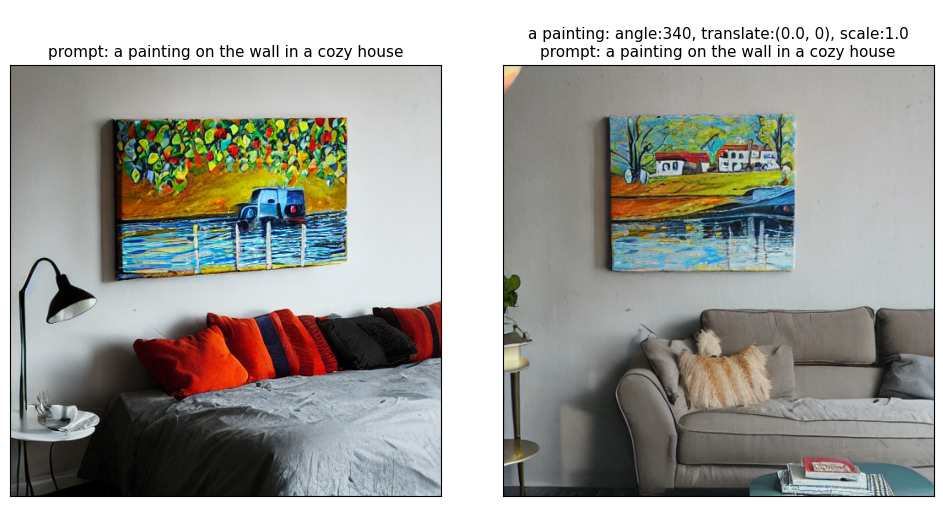}
	\end{subfigure}
    \begin{subfigure}{0.245\linewidth}
	    \includegraphics[trim=125mm 0 0 0, clip, width=\linewidth ]{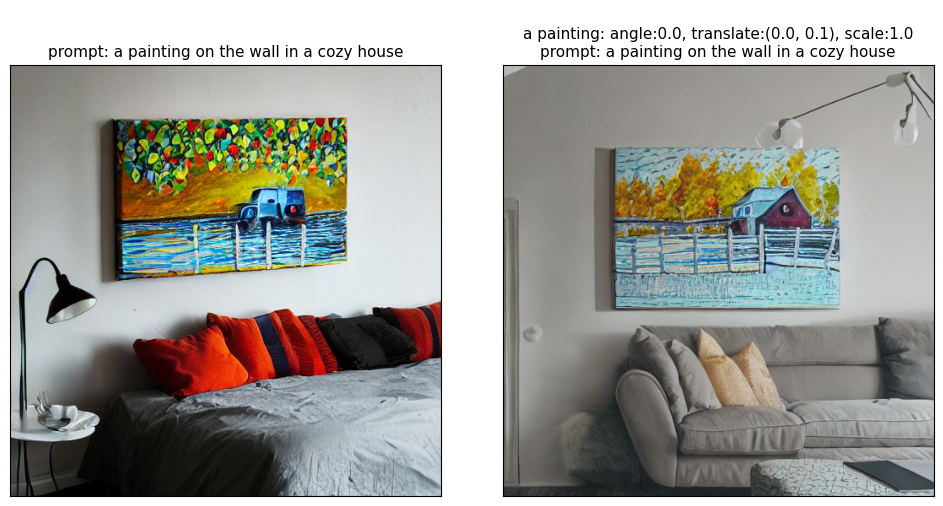}
    \end{subfigure}
	\begin{subfigure}{0.245\linewidth}
	    \includegraphics[trim=125mm 0 0 0, clip, width=\linewidth ]{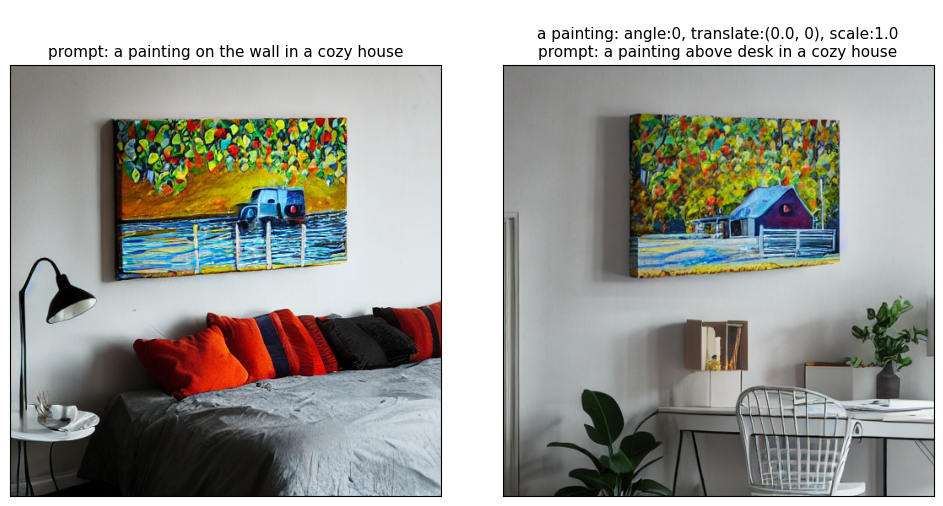}
    \end{subfigure}

   \caption{MFTF object spatial positioning (translation, rotation, scaling) and simultaneously semantic modifications.}
   \label{fig:source-target-image}
\end{figure*}

\subsubsection{Qualitative Evaluation}
The quantitative comparison of images generated using text descriptions alone for object-level layout control and those generated by MFTF model presented in the first and second rows of \cref{fig:experiment}. The source images relies solely on text descriptions fails to control object positions accurately. In contrast, MFTF model successfully positions objects by incorporating additional layout control parameters. For example, in the first pair of images, the generated image incorrectly places the ``plant" on the left side instead of the intended right side. The plant is correctly positioned on the right in MFTF model. Similarly, in the last pair of images in the second row, the generated image places ``strawberry" inside the mug on the desk instead of on the desk itself. With MFTF model, the "strawberry" is accurately positioned on the desk as intended.
The quantitative comparison of images generated with and without MFTF model for layout control is shown in the third and fourth rows of \cref{fig:experiment}. In this case, the text description does not provide specific positional control, but users may still want to adjust the positions of objects to better align with their intended layout. For example, the ``bus" is repositioned deeper into the forest, the ``moon" shrinks in size, the ``scientist" is rotated by 30 degrees, and the ``fairplace" is shown from a front view. 
\subsubsection{Quantitative Evaluation} 
Quantitative comparison of 30 image pairs generated by the Stable Diffusion model and the MFTF model is presented in \cref{tab:metrics}. In this table, the LPIPS score measures the feature-level similarity between images generated by the two models. The results indicate that feature-level similarity is well-preserved after applying layout control. Additionally, the CLIP score, also shown in \cref{tab:metrics}, evaluates the text-image alignment for both models. The minimal variation in the CLIP score demonstrates that the layout control process has negligible impact on text-image alignment, ensuring and improving consistency in semantic interpretation after layout adjustments. 
\begin{table}
\begin{tabular}{||c c c||}
    \hline
    Model &  CLIP Score & LPIPS \\ [0.5ex] 
    \hline \hline
    Stable Diffusion & 31.69 & 0.32 \\
    \hline
    MFTF & 32.26 & 0.32\\
    \hline 
\end{tabular}
\caption{Evaluation metrics comparing LPIPS and CLIP-Score for images with and without layout control.}
\label{tab:metrics}
\end{table}

\begin{figure*}
    \centering
    \begin{subfigure}{0.49\linewidth}
	    \includegraphics[width=\linewidth]{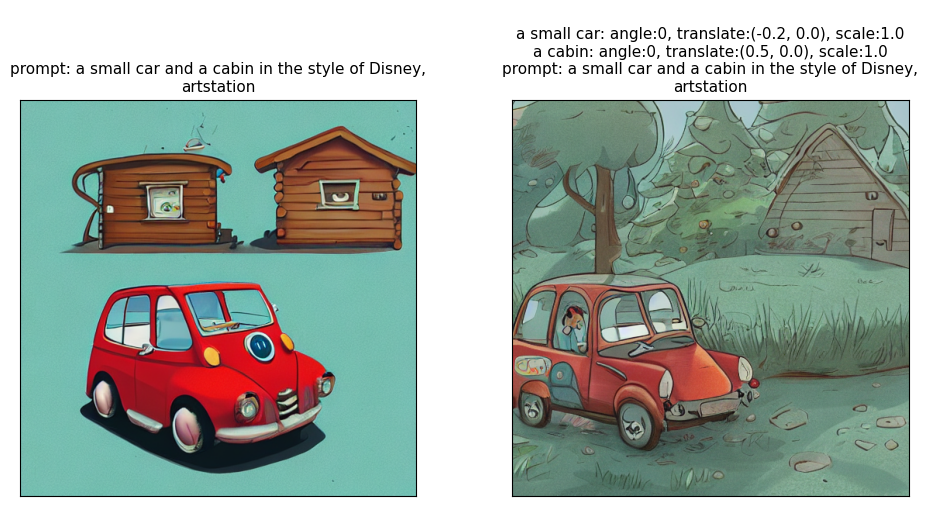}
	\end{subfigure}
	\begin{subfigure}{0.245\linewidth}
	     \includegraphics[trim=125mm 0 0 0, clip, width=\linewidth]{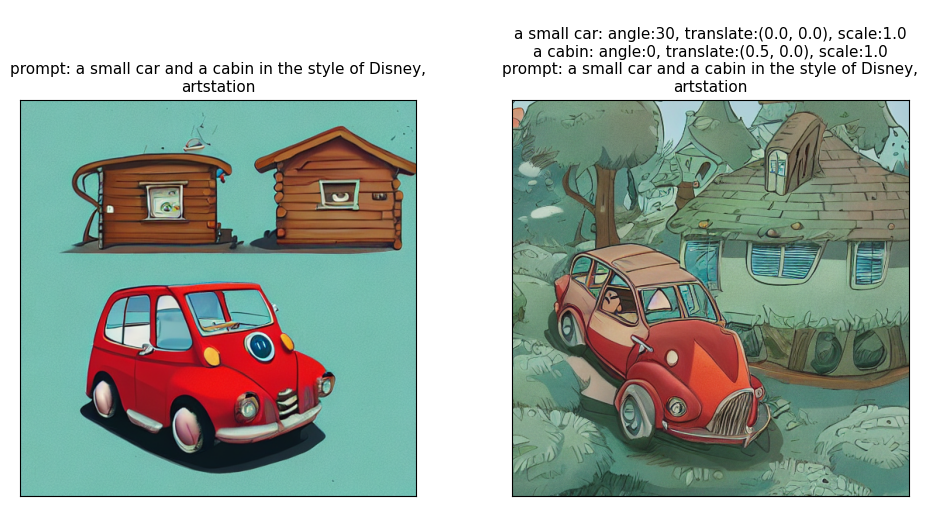}
	\end{subfigure}
    \begin{subfigure}{0.245\linewidth}
	     \includegraphics[trim=125mm 0 0 0, clip, width=\linewidth]{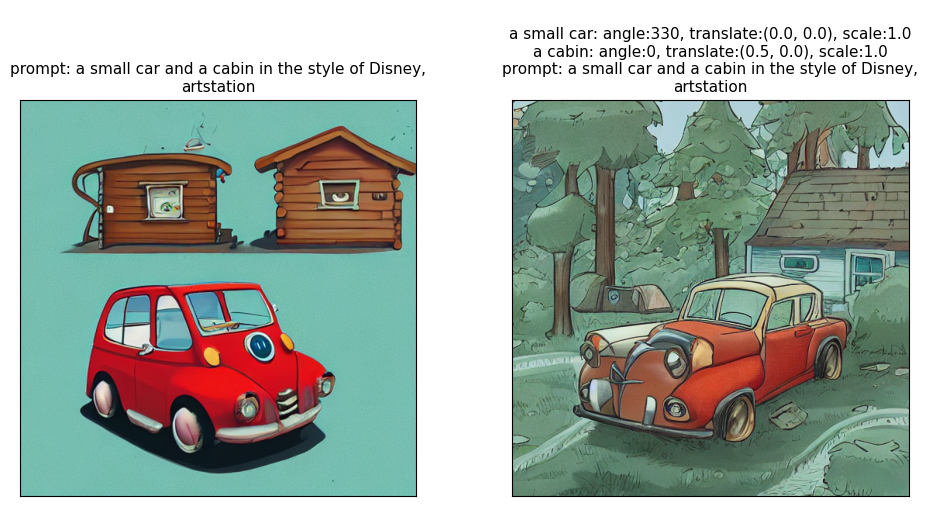}
	\end{subfigure}
    
    \vfill
    \begin{subfigure}{0.49\linewidth}
	    \includegraphics[width=\linewidth ]{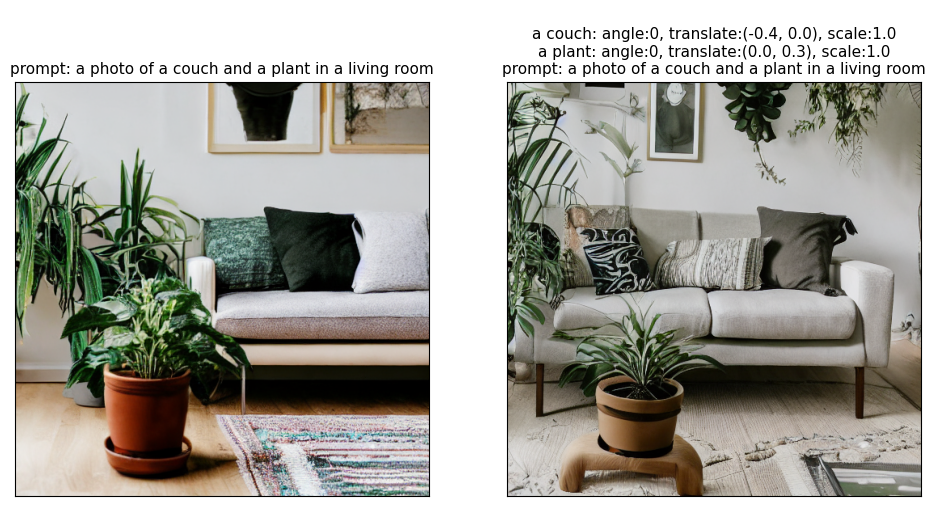}
	\end{subfigure}
    \begin{subfigure}{0.245\linewidth}
	    \includegraphics[trim=125mm 0 0 0, clip, width=\linewidth ]{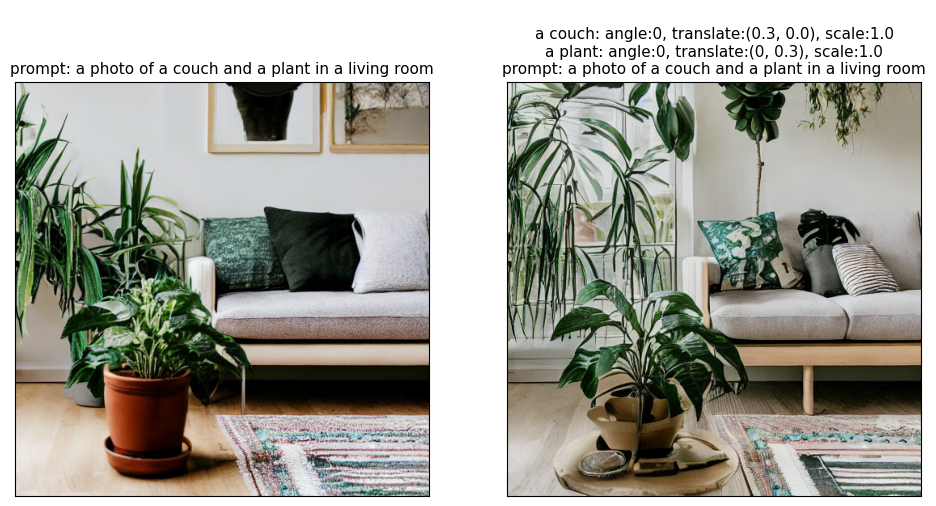}
    \end{subfigure}
	\begin{subfigure}{0.245\linewidth}
	    \includegraphics[trim=125mm 0 0 0, clip, width=\linewidth ]{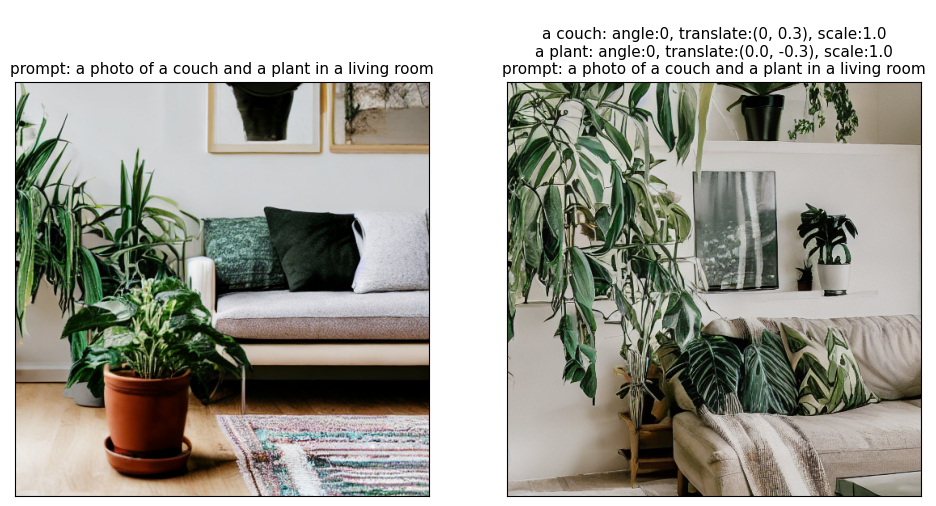}
    \end{subfigure}
    \caption{MFTF concurrently control the positions and arrangements of multiple objects within a scene, enables independent adjustments to the layout of each object while preserving overall structural coherence.}
    
   \label{fig:multi-object-control}
\end{figure*}

\subsection{Applications}
\subsubsection{Layout Control and Semantic Editing}
 The model successfully demonstrates concurrent single-object positional control and semantic editing, as illustrated in \cref{fig:source-target-image}. The first column presents images generated from the source prompts: ``a cat sitting on a chair" and "a painting on the wall in a cozy house". Following the application of layout controls $L$, the resulting images are shown in the second, third. These results highlight the ability to apply positional control, including translation, rotation, and scaling, both independently and simultaneously for target image generation. In the fourth columns of \cref{fig:source-target-image}, object-level layout control and semantic editing are applied together, with modified target prompts such as ``a dog sitting on a chair" and ``a painting above desk in a cozy house".

\subsubsection{Multiple Object Layout Control}
The model is capable of controlling the positions of multiple objects simultaneously, as demonstrated in \cref{fig:multi-object-control}. The first image in the first row, generated from the source prompt $P_s$ ``a small car and a cabin in the style of Disney, artstation", shows the initial configuration. In this case, the ``car" on the left-hand side is translated to the left, while the ``cabin" is translated to the right. 
Subsequently, the ``car" is rotated by $L=30^{\circ}$ and $L=330^{\circ}$ clockwise in the second to fourth images. The second row illustrates another example of concurrent object position control for multiple objects. 

However, the relationships between objects are difficult to decouple, and while controlling multiple object layouts concurrently is feasible, it remains constrained. The threshold $\eta$ parameter plays a critical role in determining the position of objects, especially in cases involving complex inter-object relationships. 

\subsection{Ablation Study}
\begin{figure}[h!]
   \centering
   \includegraphics[width=1\linewidth]{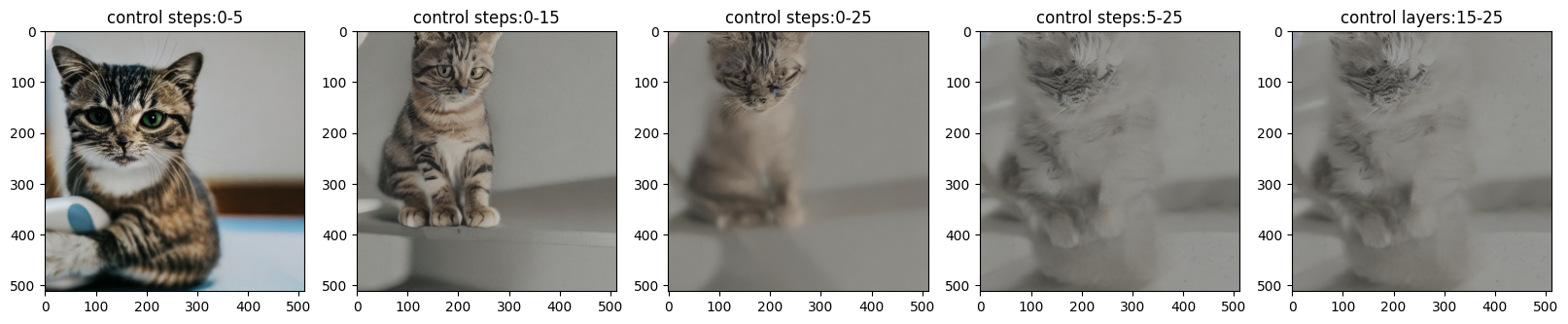}
   \caption{Applying object-level layout control across different denoising steps.}
   \label{fig:steps}
\end{figure}

\begin{figure}[h!]
  \centering
   \includegraphics[width=.8\linewidth]{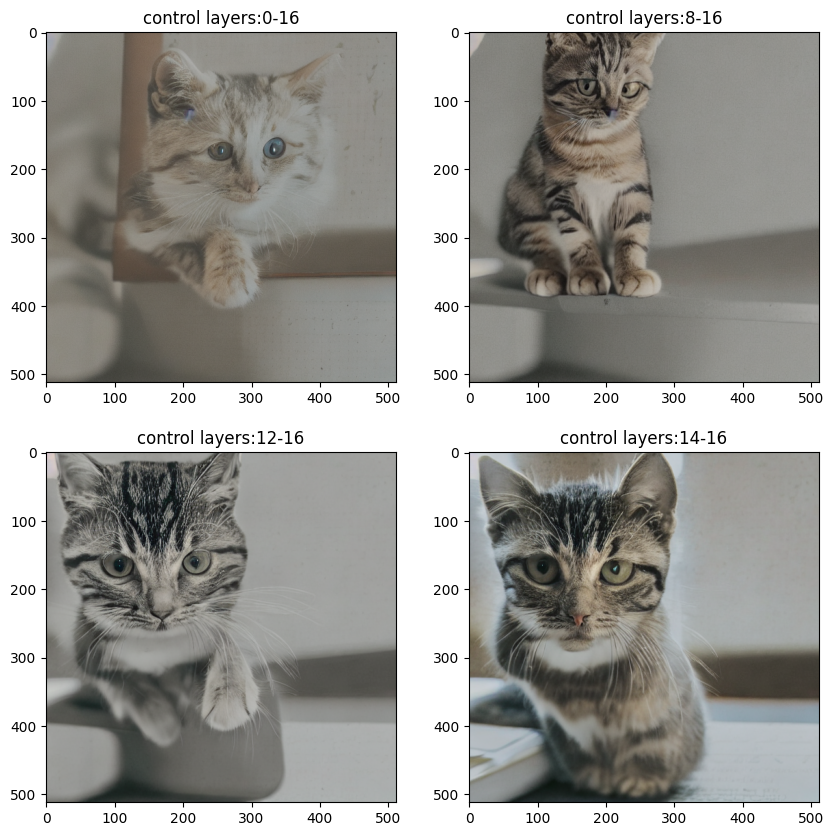}
   \caption{Applying object level layout control on different self-attention layers.}
   \label{fig:layers}
\end{figure}

The ablation study investigated the effects of layout control across denoising steps and self-attention layers. As illustrated in \cref{fig:steps}, the early denoising steps play a critical role in determining object positioning and arrangement, while the later steps refine structural details. Effective layout control requires inclusion of the initial denoising steps; omission of these steps results in a failure to achieve the desired positional control. 
The layer control ablation study, shown in \cref{fig:layers}, highlights that applying the attention mask to higher-level layers achieves effective layout control. Conversely, when the mask is applied to fewer self-attention layers, the control effectiveness decreases significantly, ultimately leading to a complete loss of positional control. These findings emphasize the importance of both step and layer selection in achieving robust layout control.

\subsection{Limitations}
A primary challenge arises from the mixed structural information present in the cross-attention layer, which makes it difficult to fully encapsulate complete structural information within a single attention mask. This limitation hinders the model's ability to fully separate objects from the background. Additionally, in multi-object position control, the model struggles with separate objects with complex interrelationships, particularly in scenarios involving tightly coupled or overlapping objects. 
\section{Conclusion}
In this paper, we have introduced MFTF, a mask-free, fine-tuning-free diffusion model for object-level layout control. It enables both single-object and multi-object layout control, supports concurrent layout manipulation and semantic editing, and facilitates text-guided image segmentation. These capabilities showcase its promising potential for advancing research and applying it could empower users to custom images generations with unprecedented ease. 
{
    \small
    \bibliographystyle{ieeenat_fullname}
    \bibliography{main}
}
\clearpage
\setcounter{page}{1}
\maketitlesupplementary

\section{Supplementary}
\label{sec:rationale}

\subsection{Modified multi-self-attention}

\subsection{Cross-attention Map}
\begin{figure}[h!]
   \includegraphics[width=1\linewidth]{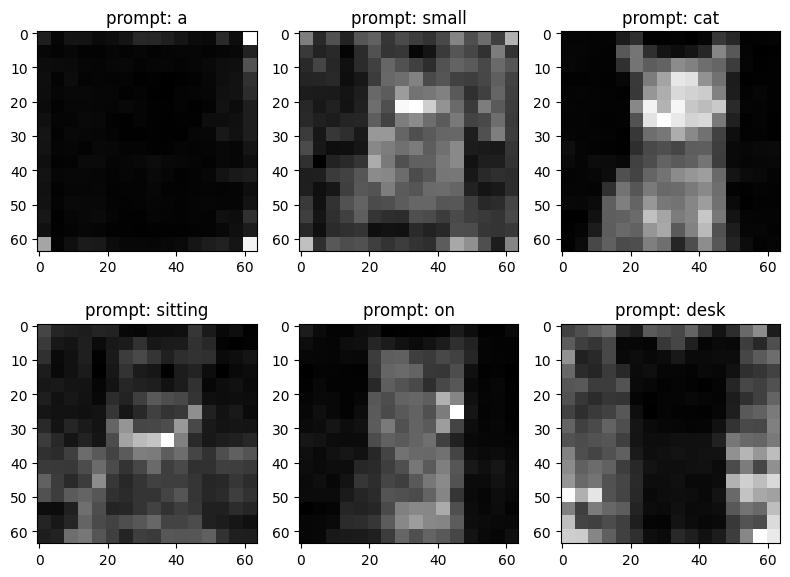}
   \caption{The prompts and their corresponding cross-attention layers $A^i_s$ are linked as follows: for each prompt, a specific cross-attention layer $A^i_s$ is extracted, which captures the interaction between the input prompt and the source features.}
   \label{fig:cross-attention-map}
\end{figure}

The cross-attention maps corresponding to individual tokens $i$ are illustrated in S\cref{fig:cross-attention-map}, which encapsulates both semantic and structural information. However, the structural information corresponding to each text token overlaps significantly, which complicates achieving precise absolute layout control. For instance, the cross-attention map in S\cref{fig:cross-attention-map} generated for the prompt ``a small cat sitting on a desk", the token ``cat" captures the majority of the structural information for the cat object. However, other tokens such as ``small", ``sitting", ``on", and ``desk" also contribute overlapping profile information related to the ``cat", thereby demonstrating the inherent redundancy in the attention map. 

\subsection{Self-attention Query}

\subsection{Text-Controlled Semantic Segmentation}
The procedure for obtaining the semantic segmentation mask is outlined as follows: first, the image is encoded into latent space $z$, with no noise added to $z$ in order to prevent introducing randomness that could distort the original object structures. The latent space $z$ is then denoised in conjunction with the prompt. Cross-attention or self-attention layers are used to extract segmentation masks for the specified tokens. While the generated images may not fully recover the original image, they effectively preserve the structural integrity of the input images for segmentation. 

\begin{figure}
  \centering
   \includegraphics[width=1\linewidth]{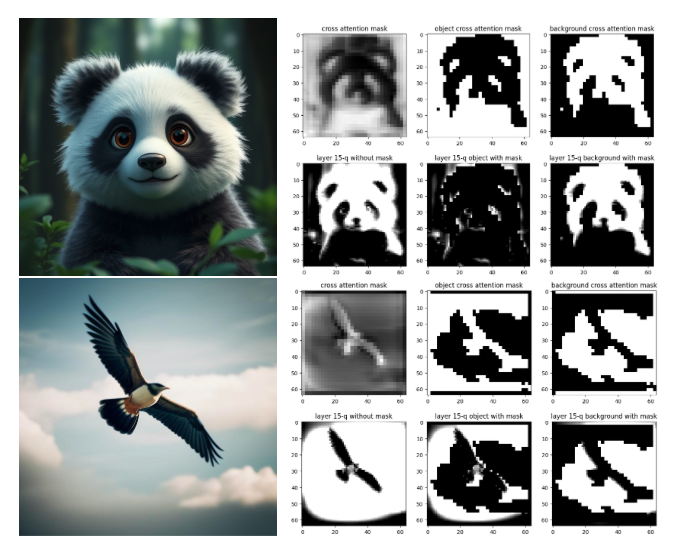}
   \caption{MFTF model enables text-controlled image segmentation through the generation of segmentation masks. Segmentation cross-attention mask $M_s$ and the masked query $Q_{ms}$ for token ``panda" and ``bird".}
   \label{fig:segmentation}
\end{figure}

The attention map $M_s$ and the masked queries, $Q_s$ or $Q_{\text{background}}$, can also be utilized for text-guided image segmentation, as illustrated in S\cref{fig:segmentation}. Here, $M_s$ serves as the mask for the token "panda", which emphasizes the object's profile, while $Q_s$ or $Q_{\text{background}}$ is used to generate detailed feature segmentation.

\subsection{Ablation Object-level Layout Control}

\end{document}